\documentclass{article}
\usepackage{spconf,amsmath,graphicx,xcolor}
\usepackage{subcaption}
\usepackage[percent]{overpic}

\usepackage[firstpage]{draftwatermark}
\SetWatermarkText{\parbox{200cm}{Copyright 2020 IEEE. Personal use of this material is permitted. Permission from IEEE must be obtained for all other uses, in any current or future media, including reprinting/republishing this material for advertising or promotional purposes, creating new collective works, for resale or redistribution to servers or lists, or reuse of any copyrighted component of this work in other works.\\ \\ \\ \\ \\ \\ \\ \\ \\ \\ \\ \\ \\ \\ \\ \\ \\ \\ \\ \\ \\ \\ \\ \\ \\ \\ \\ \\ \\ \\ \\ \\ \\ \\ \\ \\ \\ \\ \\ \\ \\}}
\SetWatermarkLightness{0.5}
\SetWatermarkScale{0.1}
\SetWatermarkAngle{0}
\SetWatermarkColor[rgb]{1,0,0}

\title{Super-resolving Commercial Satellite Imagery Using Realistic Training Data}
\name{Xiang Zhu, Hossein Talebi, Xinwei Shi, Feng Yang, Peyman Milanfar}
\address{Google Inc.}
\begin{document}
%
\maketitle
\begin{abstract}
In machine learning based single image super-resolution, the degradation model is embedded in training data generation. However, most existing satellite image super-resolution methods use a simple down-sampling model with a fixed kernel to create training images. These methods work fine on synthetic data, but do not perform well on real satellite images. We propose a realistic training data generation model for commercial satellite imagery products, which includes not only the imaging process on satellites but also the post-process on the ground. We also propose a convolutional neural network optimized for satellite images. Experiments show that the proposed training data generation model is able to improve super-resolution performance on real satellite images.
\end{abstract}
\begin{keywords}
Remote sensing, satellite imagery, super-resolution
\end{keywords}
\section{Introduction}
\label{sec:intro}

Satellite imagery has been widely used in various areas including traffic monitoring~\cite{eslami2010automatic}, land use/land cover change analysis~\cite{viana2019long}, precision agriculture~\cite{yang2012using}, natural disaster warning and management~\cite{verstappen1995aerospace}, etc. For all these applications, spatial resolution of the imagery is a key factor.

So far the lowest ground sample distance (GSD), which corresponds to the highest spatial resolution, of commercial satellite imagery products is 30cm. At the time of writing this paper, 30cm GSD is only available from satellite WorldView-3. Other sub-meter imagery products are typically of 50cm (e.g. WorldView-2, GeoEye-1, Pleiades) or 80cm (e.g. IKONOS, SkySat series) GSD. Although they are all considered very high resolution (VHR) satellites, their GSD is still not low enough for the applications mentioned above. For example, in traffic monitoring vehicles are represented by only a short number of pixels and hence the detection algorithm is very sensitive to the surrounding context~\cite{eslami2010automatic}. On the other hand, enhancing resolution via imaging hardware improvement is expensive and technically challenging~\cite{hajlaoui2010satellite}, which makes software-based image super-resolution (SR) techniques attractive in practice. 

In recent years, most successful single image SR algorithms are learning based~\cite{yang2019tmm}. The first convolutional neural network (CNN) based SR was developed by Dong~\emph{et al.}~\cite{dong2015image}, which only contains 3 convolutional layers and outputs a high-resolution (HR) image from its low-resolution (LR) input directly. Kim~\emph{et al.}~\cite{kim2016accurate} use a deep CNN with 20 layers and it is applied to generate high-frequency components (residual image) of the HR output. Then, a generative adversarial network (GAN) is introduced into the training process to make the outputs photo-realistic \cite{ledig2017photo}. Johnson~\emph{et al.}~\cite{Johnson2016eccv} proposed perceptual losses to get visually pleasing SR results. These works train models on synthetic data thus do not generalize well for real-world applications like zoom for mobile phone camera or complicated degradation. So the current trend of single image SR is to solve real-world problems~\cite{Chen_2019_CVPR} or consider more sophisticated degradations~\cite{Zhang_2018_CVPR}.

There are also several SR approaches specifically designed for satellite imagery. Some of them are basically implementation of existing learning based approaches with some modifications, and use satellite images as training data \cite{liebel2016single,pouliot2018landsat}. GANs are also used to improve low-resolution texture restoration~\cite{reshad2019deep,bosch2018super}. Jiang~\emph{et al.}~\cite{jiang2019edge} developed a method that combines residual image enhancement and GAN together. However, none of them paid much attention to the actual image degradation model of satellite images. For learning based SR, the degradation model is embedded in training data, especially in the LR image simulation. In \cite{liebel2016single,reshad2019deep} the LR training data is created via simple down-scaling of HR images, which basically assumes the following degradation model:
\begin{equation}
    \mathbf{y}=D(\mathbf{h}*\mathbf{z}),\label{eq:1}
\end{equation}
where $\mathbf{y}$ is the observed LR image, and $\mathbf{z}$ denotes the latent HR image. $\mathbf{h}$ represents a fixed blur kernel (e.g. bicubic kernel), $D(.)$ represents the down-sampling operation, and $*$ is a convolution operator. This model has been widely used in many natural image SR approaches, which treat SR as a non-blind deconvolution problem.

\begin{figure}[t!]
    \centering
    \begin{subfigure}[t]{0.12\textwidth}
        \centering
        \includegraphics[bb={57 30 330 300},clip,width=\textwidth]{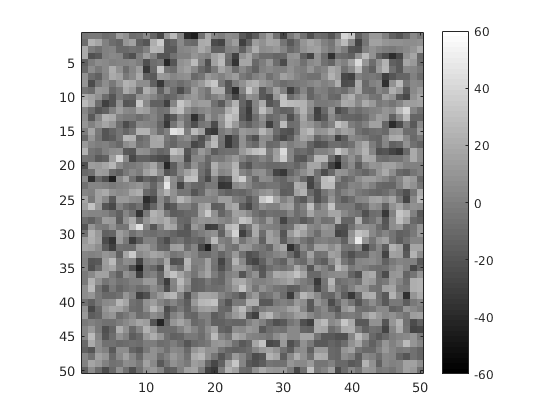}
        \caption{}
    \end{subfigure}
    \begin{subfigure}[t]{0.12\textwidth}
        \centering
        \includegraphics[bb={62 30 330 300},clip,width=\textwidth]{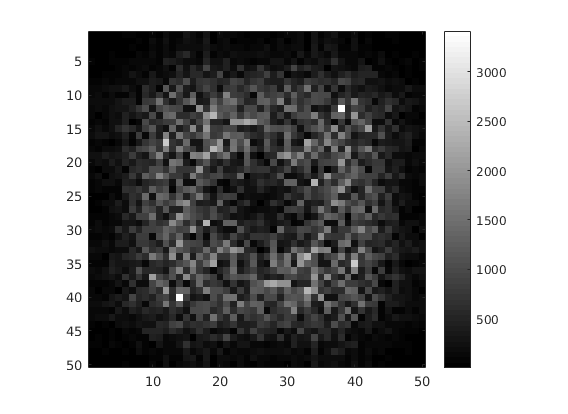}
        \caption{}
    \end{subfigure}
    \begin{subfigure}[t]{0.1\textwidth}
        \centering
        \includegraphics[bb={64 30 330 300},clip,width=\textwidth]{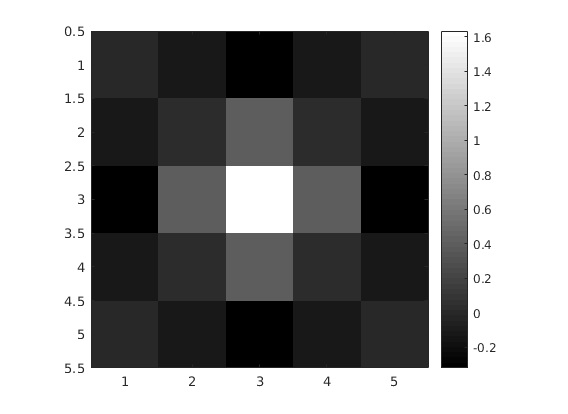}
        \caption{}
    \end{subfigure}
    \begin{subfigure}[t]{0.12\textwidth}
        \centering
        \includegraphics[bb={62 30 330 300},clip,width=\textwidth]{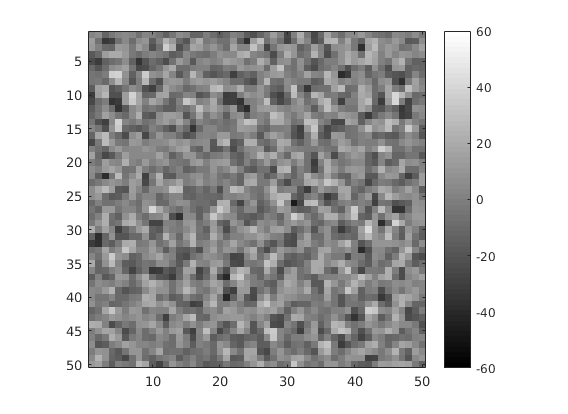}
        \caption{}
    \end{subfigure}
    \caption{Satellite image noise analysis. (a) Noise sample extracted from a Pleiades image. (b) FFT of (a). (c) Blur kernel estimated from (b). (d) Noise simulated by convolving WGN with (c).}
    \label{fig:sat_noise}
\end{figure}

Unfortunately, the model in \eqref{eq:1} is not realistic for commercial satellite images. First of all it lacks noise, and most satellite images are highly noisy. Secondly, the point spread function (PSF) in the satellite imaging system needs to be considered. Bicubic kernel is not a good approximation of the real PSFs. In fact, PSFs of most time delay and integration (TDI) sensors on satellites are spatially variant due to the imaging hardware limitation \cite{hajlaoui2010satellite}. There also exists motion blur caused by satellite movement or sensor scanning. In other words, satellite image super-resolution is more like a blind deconvolution problem. Thirdly, it forgets that most commercial satellite image products are processed after imaging by their providers, and the process usually includes resampling, which further blurs images, and changes the distribution of noise so that it can no longer be treated as white Gaussian noise (WGN). All these factors need to be considered in training data generation.

In this paper, we propose a realistic training data generation model with spatially variant PSFs based on our analysis of commercial satellite images. We also proposed a CNN-based super-resolution model which is able to handle variant PSFs and different degrees of aliasing. We use a residual CNN architecture similar to \cite{ledig2017photo} with some modifications to make the model more efficient. In the experiments section we will show its performance on real satellite imagery products.

\section{Methodology}
\label{sec:method}

\subsection{Degradation Model}
\label{ssec:deform}

We use the following model to describe the degradation process of commercial satellite images.
\begin{equation}
    \mathbf{y}=R(\mathbf{h}_{p}*(D(\mathbf{h}_{o}*\mathbf{z}) + \mathbf{n})).\label{eq:2}
\end{equation}
$\mathbf{h}_o$ is the PSF of the imaging system, and it is assumed to be variant spatially and over images, though it should vary slowly and can be viewed invariant in small local regions of $\mathbf{z}$ ~\cite{hajlaoui2010satellite}. $\mathbf{n}$ denotes additive noise from the imaging system, which can be approximated as WGN. $\mathbf{h}_{p}$ represents a resampling kernel introduced in the post-process on the ground, and $R(.)$ denotes a resampling operation. Resampling is needed in the post-process in order to align pixels from multiple channels to a target coordinate grid, which is associated with the image's camera model. Locally $R(.)$ can be approximated as spatial shifting.

By merging the post-process and imaging model together we can rewrite \eqref{eq:2} as
\begin{equation}
    \mathbf{y}=D(\mathbf{h}_{m}*\dot{\mathbf{z}}) + \tilde{\mathbf{n}},\label{eq:3}
\end{equation}
where $\dot{\mathbf{z}}$ denotes the HR image on the up-sampled target coordinate grid, and $\mathbf{h}_{m}$ represents a kernel mixing the effect of $\mathbf{h}_{p}$ and $\mathbf{h}_{o}$ together. $\mathbf{h}_{m}$ is spatially varying within and across images from a same satellite sensor, but in a local image area it can be treated as invariant. $\tilde{\mathbf{n}}$ denotes the final noise effect, and can be treated as WGN convolved by $\mathbf{h}_{p}$.

\begin{figure}[t!]
    \centering
    \begin{subfigure}[t]{0.15\textwidth}
        \centering
        \includegraphics[width=\textwidth]{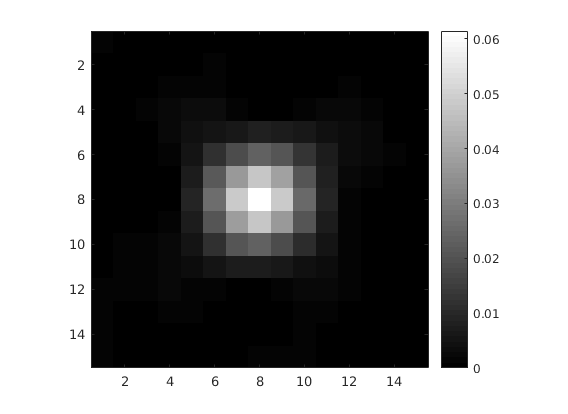}
    \end{subfigure}
    \begin{subfigure}[t]{0.15\textwidth}
        \centering
        \includegraphics[width=\textwidth]{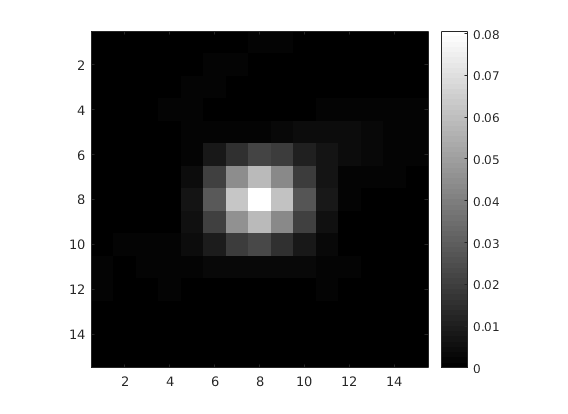}
    \end{subfigure}
    \begin{subfigure}[t]{0.15\textwidth}
        \centering
        \includegraphics[width=\textwidth]{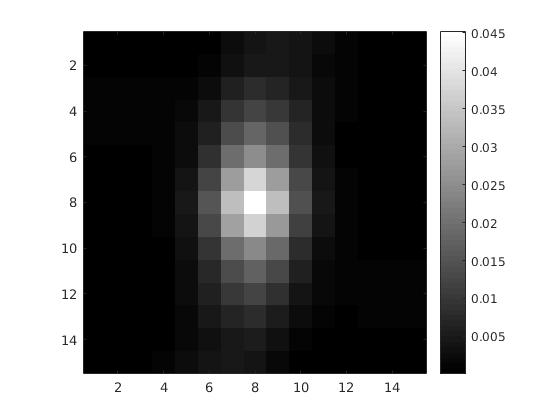}
    \end{subfigure}
    \caption{PSF samples estimated from GeoEye-1 images.}
    \label{fig:sat_psf}
\end{figure}

To verify the degradation model in \eqref{eq:3} we analyzed some satellite images. Fig.~\ref{fig:sat_noise} (a) shows a noise sample extracted from a flat area of a 16-bit Pleiades image. Its 2-D Fourier transformation in (b) indicates that it is colored. Our analysis further shows that such noise can be simulated by convolving WGN with an estimated kernel (See Fig.~\ref{fig:sat_noise} (c) and (d)), which matches the proposed model in \eqref{eq:2}.

PSFs are also analyzed. We estimated PSFs from local image regions via a shock filter based method similar to ~\cite{money2008total}. Three PSFs estimated from GeoEye-1 images are shown in Fig.~\ref{fig:sat_psf}, and it turns out they are variant not only in their spread but also in their shape. Mild motion blur is observed (see the 3rd PSF in Fig.~\ref{fig:sat_psf}), and since it is along column direction it could be introduced by sensor scanning.

\subsection{Training Data Generation}
\label{ssec:data}

Instead of using satellite images with relatively low SNR and potential motion blur to generate training data as most existing methods did, we choose Google owned aerial images as the source to create synthetic LR images. LR images are generated via \eqref{eq:3}. Noise $\tilde{\mathbf{n}}$ is simulated by convolving WGN with an estimated kernel (such as Fig.~\ref{fig:sat_noise}(c)). Each satellite product type has a corresponding noise kernel. $\mathbf{h}_{m}$ is simulated via a 2-D elliptical Gaussian mixture model:
\begin{equation}
    \mathbf{h}_{m}(x_{1},x_{2})=\frac{1}{Z}\sum_{i} \alpha_{i} \cdot exp \left( -\left(\frac{x_{1}^2}{2\sigma_{i,1}^2} + \frac{x_{2}^2}{2\sigma_{i,2}^2} \right) \right),\label{eq:4}
\end{equation}
where each $(\sigma_{i,1}, \sigma_{i,2})$ controls the shape of the $i$-th Gaussian component, and $\{\alpha_{i}\}$ denotes its contribution. $Z$ is the overall normalization factor. Several PSFs with various shapes are estimated from real images, and for each PSF a set of $\{\alpha_{i}\}$ is derived via least square fitting. When generating a LR image, a set of $\{\alpha_{i}\}$ is randomly selected. We then add a little noise to the selected $\{\alpha_{i}\}$ to further varies the shape of the synthetic PSF. We also varies the down-scaling factor within a small range to simulate the blur and aliasing variation we observed in real satellite images.

\subsection{Network Architecture}
\label{ssec:net}

\begin{figure}[!t]
\vspace{-0 mm}
\begin{center}
\includegraphics*[viewport=1 1 280 260, scale=0.85]{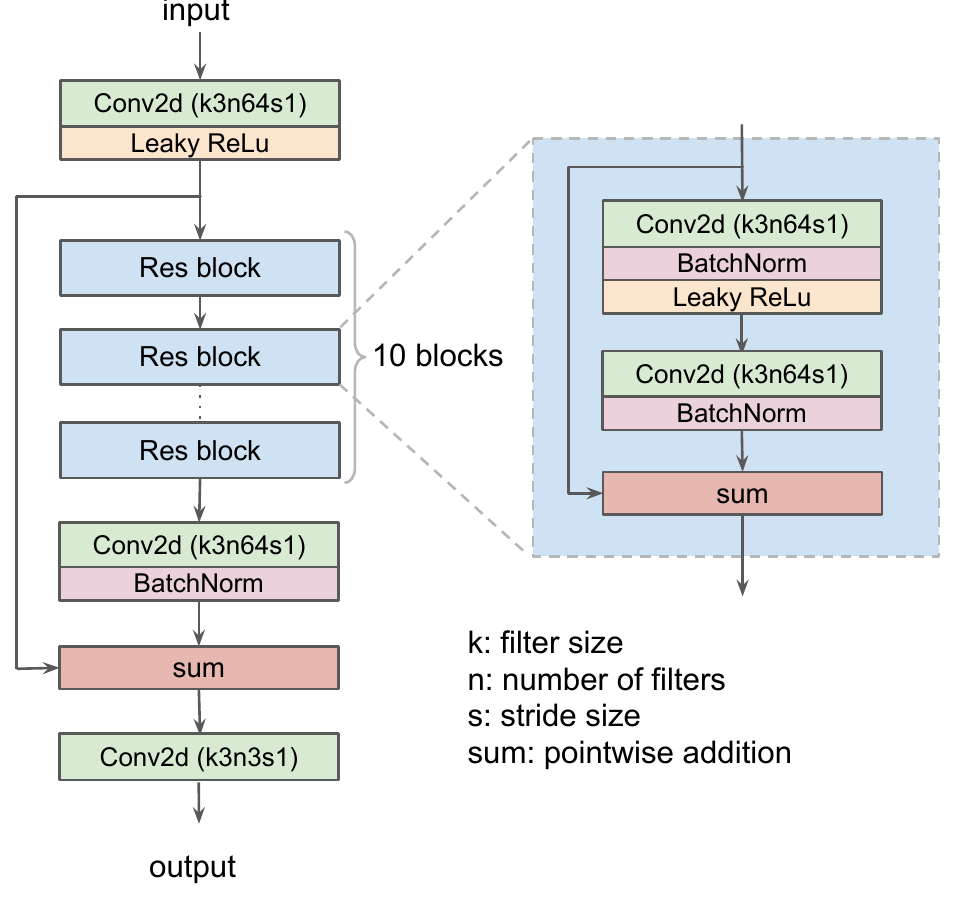}
\end{center}
\vspace{-6 mm}
{\caption{Our residual-based CNN model.} \label{fig:residual_cnn_architecture}}
\vspace{-4 mm}
\end{figure}

Our CNN architecture is shown in Fig.~\ref{fig:residual_cnn_architecture}. This CNN is similar to the model used by Ledig et al.~\cite{ledig2017photo}. All intermediate layers have $64$ convolutional filters of size $3\times3$, followed by Leaky ReLU activations with slope of $0.2$ for negative values. We use $10$ identical residual blocks, that account for a large filter footprint of size $41\times41$. This allows for a more effective model, specially when dealing with large PSFs from satellite images.

Input and output of our network are 3-channel pan-sharpened RGB images in linear space. We also tried YCbCr color space, where only Y channel was fed to the network. It turns out that using RGB images in our training leads to better noise suppression.

Note that input image needs to be upscaled to the target HR image size with bicubic interpolation before being fed to the CNN model. This enables the network to handle flexible upscaling factors, and hence to accurately adjust the GSD of its output images. 

\subsection{Training Loss}
\label{ssec:optim}

Similar to the framework of \cite{Talebi2018noref,talebi2018learned}, our training loss has two terms: a fidelity loss, and a perceptual loss:

\begin{equation}
\label{eqn:loss}
l_{\textbf{W}} = f(\textbf{z}, c_{\textbf{W}}(\textbf{y}^\prime)) + \gamma q(c_{\textbf{W}}(\textbf{y}^\prime)).
\end{equation}

The fidelity loss $f(.)$ enforces closeness of the output image $c_{\textbf{W}}(\textbf{y}^\prime)$ to the ground truth high-resolution image $\textbf{z}$. $c_{\textbf{W}}(.)$ denotes our CNN model with trainable weights ${\textbf{W}}$. $\textbf{y}^\prime$ is the bicubic upscaled input image. The fidelity loss $f(.)$  is a pseudo-Huber function~\cite{pseudoHuber} that is a smooth approximation of the Huber loss function which combines the $L_2$ squared-loss for small differences and the $L_1$ absolute-loss for large ones, while being strongly convex.

The perceptual loss $q(.)$ is a neural network trained for no-reference image quality assessment~\cite{Talebi2018noref}. This network is differentiable, and can be plugged into our training framework. Function $q(.)$ is inversely related to predicted quality score as $q(\textbf{x}) = 10 - s(\textbf{x})$, where $s(\textbf{x})$ is the predicted positive score for image $\textbf{x}$ with $10$ as its maximum possible value. We observed that setting an appropriate weight $\gamma$ can improve upon the fidelity loss by adding more fine-grained details into the output image. We fix $\gamma$ as $0.001$ during training.

\begin{figure}[b!]
    \centering
    \begin{subfigure}[t]{0.15\textwidth}
        \centering
        \includegraphics[bb={150 170 250 240},clip,width=\textwidth]{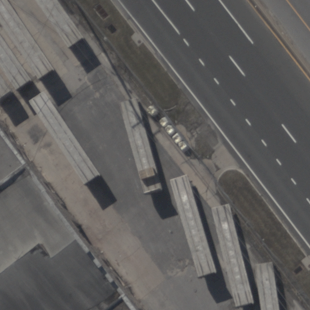}
    \end{subfigure}
    \begin{subfigure}[t]{0.15\textwidth}
        \centering
        \includegraphics[bb={30 100 130 170},clip,width=\textwidth]{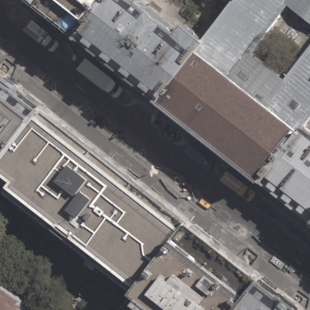}
    \end{subfigure}
    \begin{subfigure}[t]{0.15\textwidth}
        \centering
        \includegraphics[bb={150 220 250 290},clip,width=\textwidth]{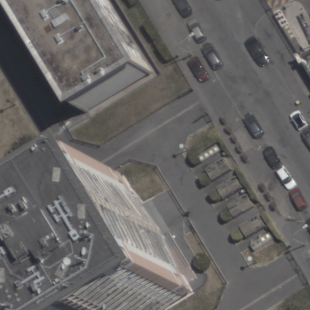}
    \end{subfigure}
    \vspace{0.5mm}
    
    \begin{subfigure}[t]{0.15\textwidth}
        \centering
        \begin{overpic}[bb={150 170 250 240},clip,width=\textwidth]{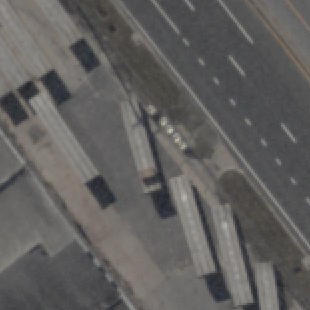}
            \put(70,0){\includegraphics[bb={70 35 320 285},clip,width=0.3\textwidth]{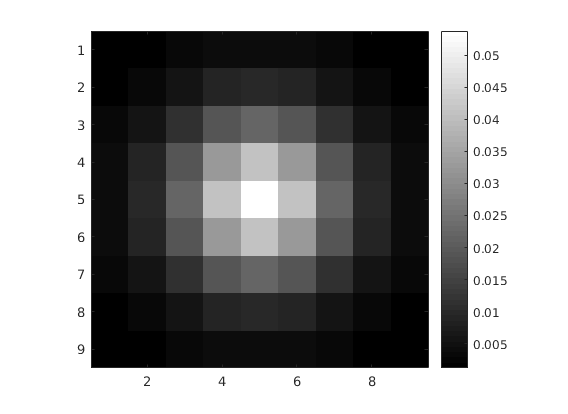}}
        \end{overpic}
    \end{subfigure}
    \begin{subfigure}[t]{0.15\textwidth}
        \centering
        \begin{overpic}[bb={30 100 130 170},clip,width=\textwidth]{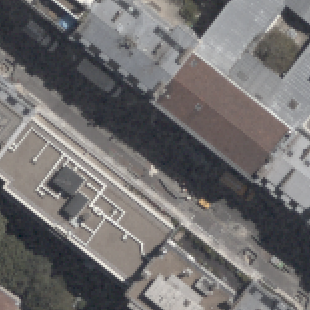}
            \put(70,0){\includegraphics[bb={70 35 320 285},clip,width=0.3\textwidth]{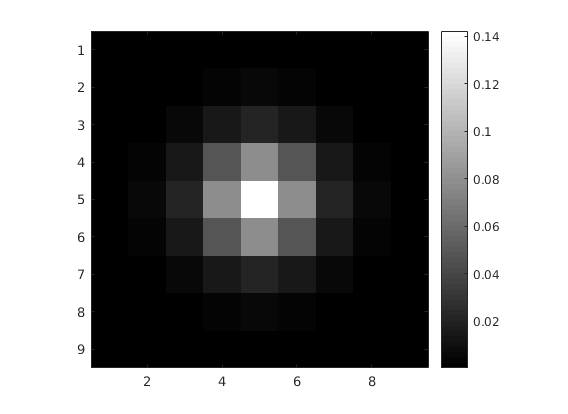}}
        \end{overpic}
    \end{subfigure}
    \begin{subfigure}[t]{0.15\textwidth}
        \centering
        \begin{overpic}[bb={150 220 250 290},clip,width=\textwidth]{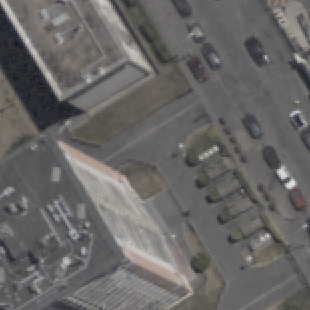}
            \put(70,0){\includegraphics[bb={70 35 320 285},clip,width=0.3\textwidth]{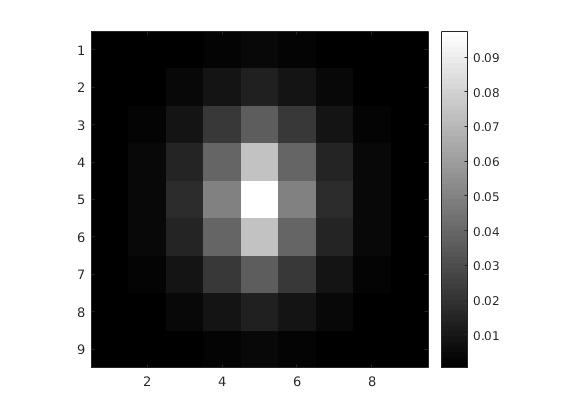}}
        \end{overpic}
    \end{subfigure}
    \vspace{0.2mm}
    
    \begin{subfigure}[t]{0.15\textwidth}
        \centering
        \includegraphics[bb={150 170 250 240},clip,width=\textwidth]{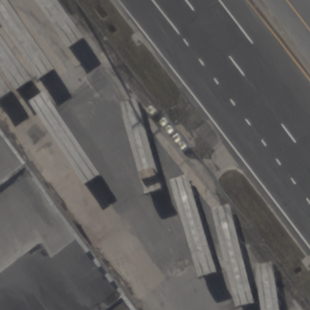}
    \end{subfigure}
    \begin{subfigure}[t]{0.15\textwidth}
        \centering
        \includegraphics[bb={30 100 130 170},clip,width=\textwidth]{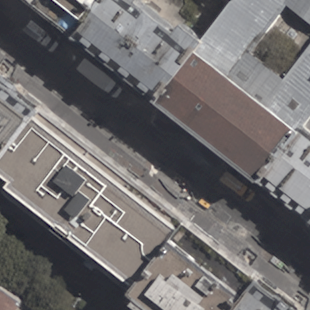}
    \end{subfigure}
    \begin{subfigure}[t]{0.15\textwidth}
        \centering
        \includegraphics[bb={150 220 250 290},clip,width=\textwidth]{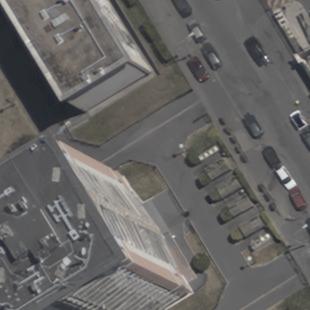}
    \end{subfigure}
    \vspace{-0.5mm}
    \caption{Synthetic image super-resolution results. 1st row: original HR images. 2nd row: synthetic input LR images with their PSF. 3rd row: SR outputs.}
    \label{fig:synthetic}
\end{figure}

\begin{figure*}[t!]
    \centering
    \begin{subfigure}[t]{0.24\textwidth}
        \centering
        \includegraphics[bb={100 300 300 420},clip,width=\textwidth]{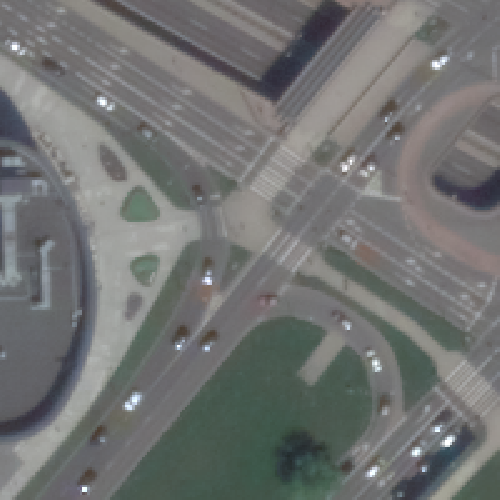}
    \end{subfigure}
    \begin{subfigure}[t]{0.24\textwidth}
        \centering
        \includegraphics[bb={0 100 200 220},clip,width=\textwidth]{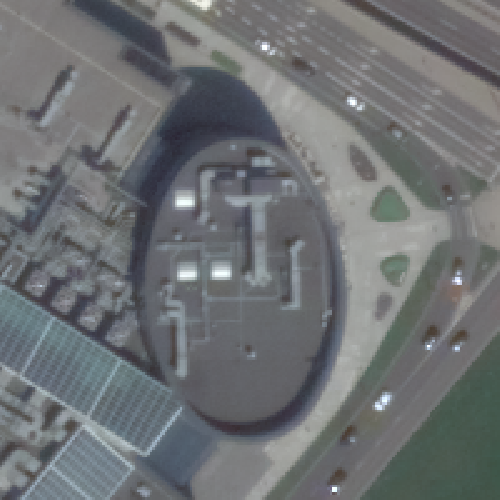}
    \end{subfigure}
    \begin{subfigure}[t]{0.24\textwidth}
        \centering
        \includegraphics[bb={150 100 350 220},clip,width=\textwidth]{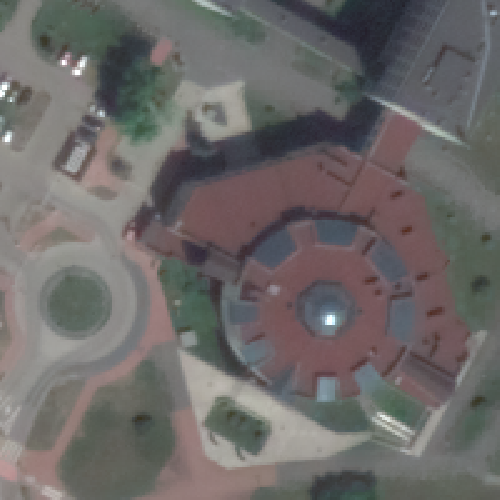}
    \end{subfigure}
    \begin{subfigure}[t]{0.24\textwidth}
        \centering
        \includegraphics[bb={150 330 350 450},clip,width=\textwidth]{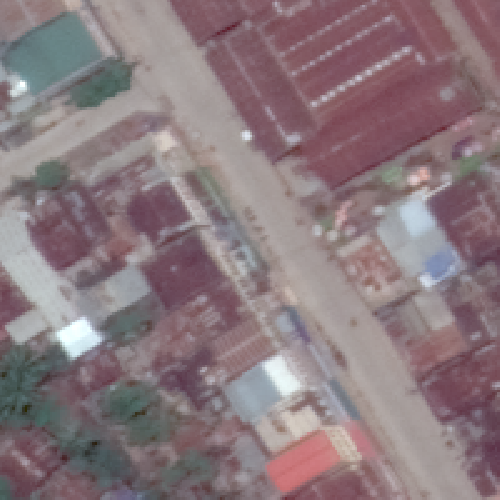}
    \end{subfigure}
    
    \begin{subfigure}[t]{0.24\textwidth}
        \centering
        \includegraphics[bb={100 300 300 420},clip,width=\textwidth]{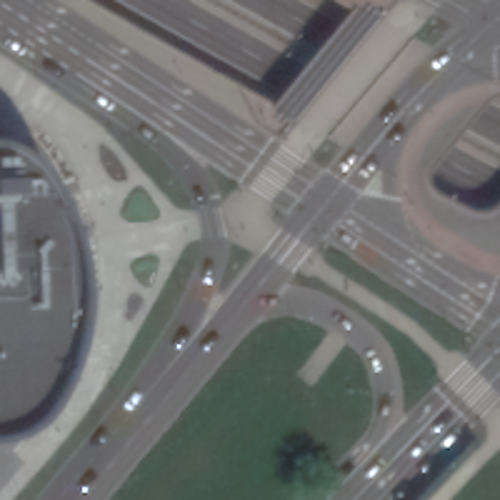}
    \end{subfigure}
    \begin{subfigure}[t]{0.24\textwidth}
        \centering
        \includegraphics[bb={0 100 200 220},clip,width=\textwidth]{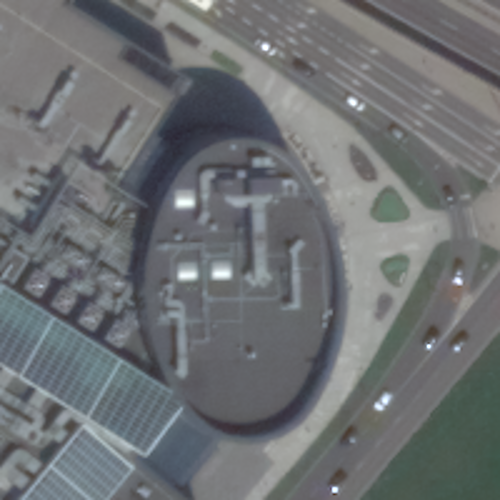}
    \end{subfigure}
    \begin{subfigure}[t]{0.24\textwidth}
        \centering
        \includegraphics[bb={150 100 350 220},clip,width=\textwidth]{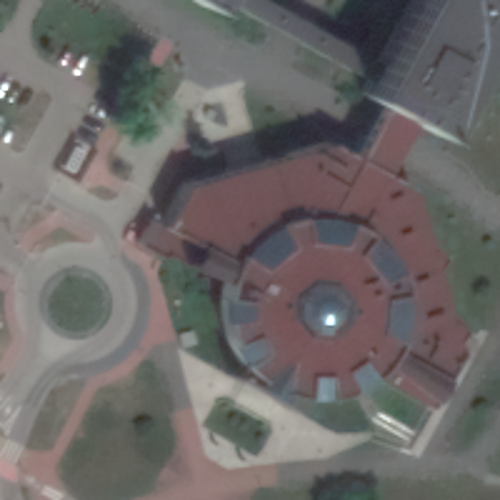}
    \end{subfigure}
    \begin{subfigure}[t]{0.24\textwidth}
        \centering
        \includegraphics[bb={150 330 350 450},clip,width=\textwidth]{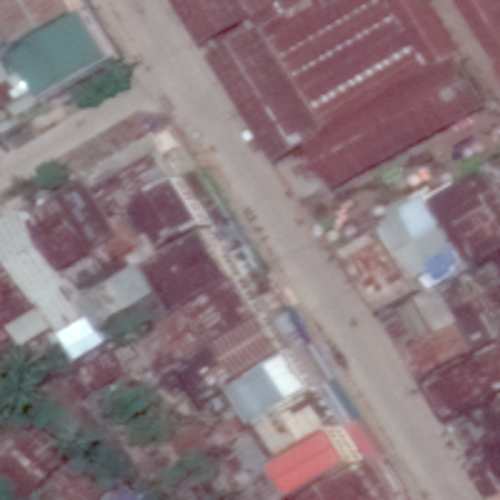}
    \end{subfigure}
    
    \begin{subfigure}[t]{0.24\textwidth}
        \centering
        \includegraphics[bb={100 300 300 420},clip,width=\textwidth]{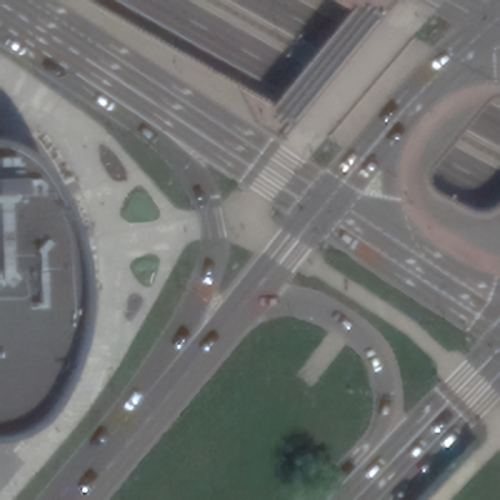}
    \end{subfigure}
    \begin{subfigure}[t]{0.24\textwidth}
        \centering
        \includegraphics[bb={0 100 200 220},clip,width=\textwidth]{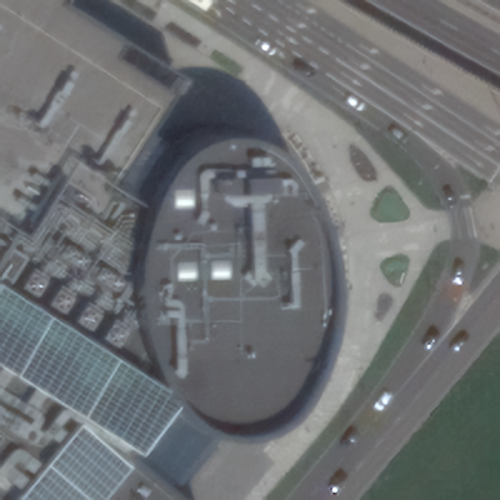}
    \end{subfigure}
    \begin{subfigure}[t]{0.24\textwidth}
        \centering
        \includegraphics[bb={150 100 350 220},clip,width=\textwidth]{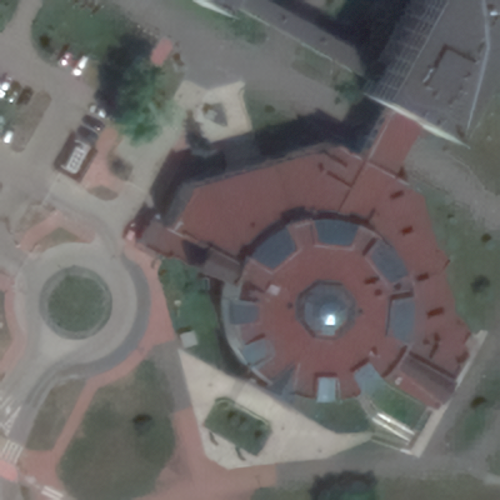}
    \end{subfigure}
    \begin{subfigure}[t]{0.24\textwidth}
        \centering
        \includegraphics[bb={150 330 350 450},clip,width=\textwidth]{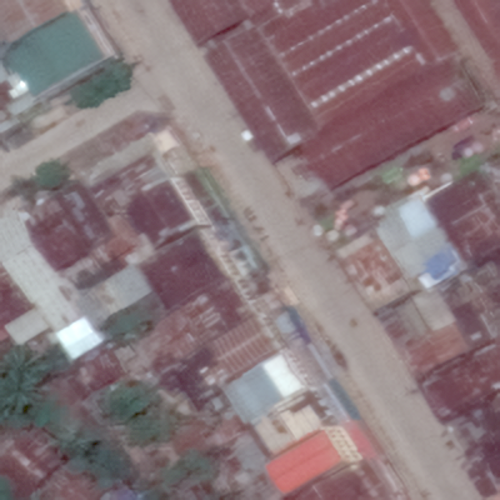}
    \end{subfigure}
    
    \begin{subfigure}[t]{0.24\textwidth}
        \centering
        \includegraphics[bb={100 300 300 420},clip,width=\textwidth]{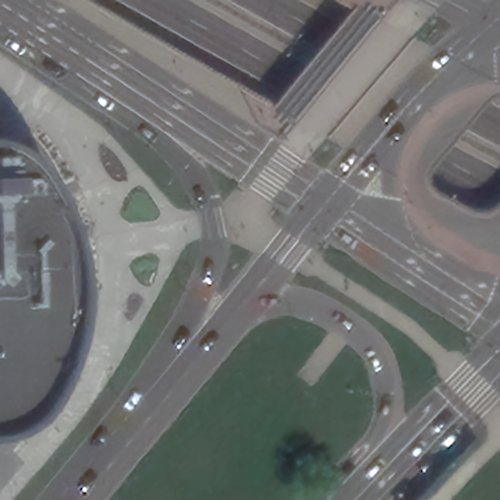}
    \end{subfigure}
    \begin{subfigure}[t]{0.24\textwidth}
        \centering
        \includegraphics[bb={0 100 200 220},clip,width=\textwidth]{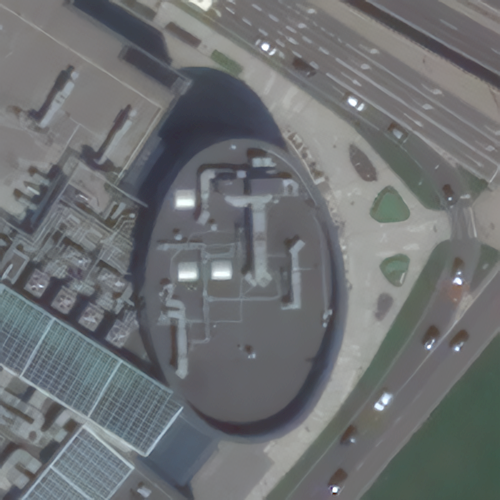}
    \end{subfigure}
    \begin{subfigure}[t]{0.24\textwidth}
        \centering
        \includegraphics[bb={150 100 350 220},clip,width=\textwidth]{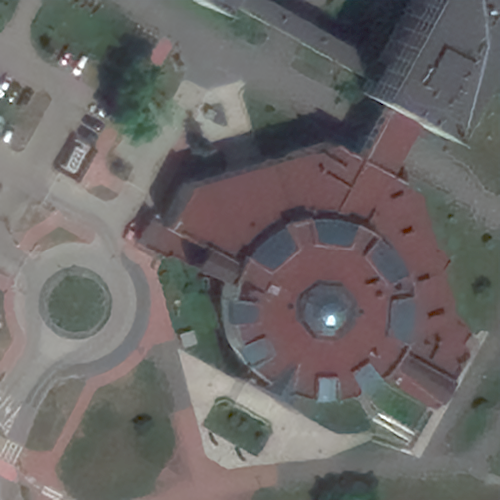}
    \end{subfigure}
    \begin{subfigure}[t]{0.24\textwidth}
        \centering
        \includegraphics[bb={150 330 350 450},clip,width=\textwidth]{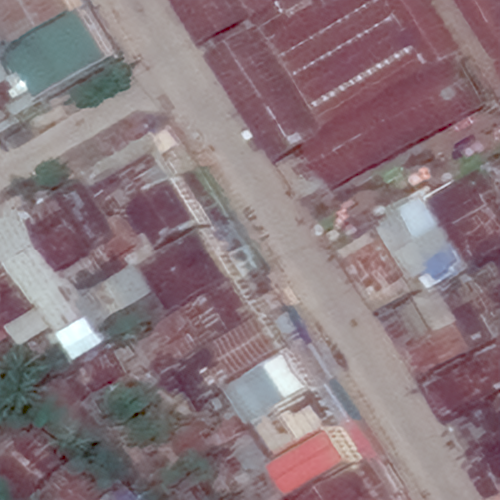}
    \end{subfigure}

    \caption{Real 50cm GeoEye-1 image\protect\footnotemark SR results. GSD is decreased from 50cm to 25cm. 1st row: LR inputs. 2nd row: bicubic interpolation. 3rd row: SR with bicubic down-scaling model. 4th row: proposed SR outputs.}
    \label{fig:geoeye1}
\end{figure*}

\begin{figure}[b!]
    \centering
    \includegraphics[bb={105 235 500 535},clip,width=0.5\textwidth]{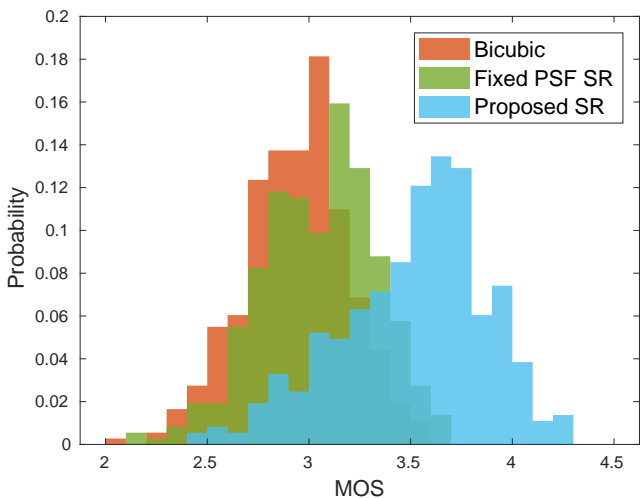}
    \caption{MOS histograms of bicubic interpolation, SR with fixed kernel bicubic down-scaling model, SR with proposed degradation model. }
    \label{fig:histo}
\end{figure}

\section{Experimental Results}
\label{sec:experi}
Training data is generated via the process described in Section~\ref{ssec:deform}. Aerial images are in linear color space, and they are divided into patches of size $310\times 310$. Each satellite sensor has a corresponding training dataset with the estimated noise kernel and PSF parameters, and one set contains 20,000 patches. 

We first show the results with synthetic images generated through the same process as the training data. Three examples with different blur and aliasing conditions are given in Fig. \ref{fig:synthetic}. All the restored images look close to their HR version. This illustrates our neural network's ability to blindly remove blur (including motion blur) and aliasing artifacts.

The trained neural network is then applied to real satellite images. Results from several 50cm GeoEye-1 images are shown in Fig.~\ref{fig:geoeye1}. The target GSD is 25cm, which is lower than any existing commercial satellite. A lot of useful high-frequency image components including vehicle details, pedestrian crossing lines, solar panel cells, and pipe infrastructure on building roofs have been successfully restored. Results from the same network but trained with the bicubic down-scaling model~\eqref{eq:1} are also given as comparison. Though their image sharpness also get improved compared with bicubic interpolation, the improvement is much limited.

To quantitatively evaluate the proposed training data generation model's performance, we randomly sampled 364 GeoEye-1 images (of size $1000\times 1000$), which are then up-scaled ($\times 2$) using bicubic interpolation, SR with generation model ~\eqref{eq:1}, and SR with the proposed model~\eqref{eq:3} respectively. We sent the images to ordinary viewers for visual quality evaluation, and the mean opinion score (MOS) of each image was then derived from 30 responses. The MOS histograms of three methods are shown in Fig.~\ref{fig:histo}, where the proposed model significantly outperforms the other two.

\footnotetext{Image \copyright 2020 Maxar Technologies.}

\section{Conclusion}
\label{sec:conclude}

We proposed a realistic SR training data generation model for commercial satellite images. The model includes not only the imaging process on satellites but also the post-process on the ground. A SR neural network is also developed to apply this model. Experiments show that our method is able to recover fine details from real satellite images.

So far the parameters of the training data generation model need to be manually estimated and tuned from satellite image samples. In the future, we will explore to use GAN to automatically generate the parameters given HR source images and the target LR image samples.

\bibliographystyle{IEEEbib}
\bibliography{refs}
\end{document}